\documentclass{article}
\usepackage{spconf,amsmath,graphicx,hyperref}
\usepackage{amsmath}
\usepackage{amsfonts}
\usepackage{pifont}   
\newcommand{\cmark}{\ding{51}}%
\usepackage{multirow}

\title{ReLo-IRR: Reflection-Guided LoRA Framework for Image Reflection Removal}
\name{Chaoqun Wang$^{1}$, Yuehuan Wei$^{1}$, Haoxiang Cao$^{1}$, Shaobo Min$^{2}$}
\address{
    $^{1}$School of Artificial Intelligence, South China Normal University, Guangzhou, China\\
    $^{2}$University of Science and Technology of China, Hefei, China
}

\begin{document}
\ninept
\maketitle
\begin{abstract}
Single-image reflection removal (SIRR) aims to recover the clean transmission layer from a reflection-contaminated image.
Although recent methods achieve promising results with large diffusion models, they rely on image-agnostic adaptation strategies, \emph{e.g.}, fine-tuning or ControlNet, that enforce uniform suppression regardless of reflection severity. As a result, heavy reflections often leave residuals, while weak ones suffer from detail loss. To this end, we propose ReLo-IRR, a reflection-guided LoRA framework built upon the rectified flow model. First, a lightweight estimator is designed to predict the reflection strength descriptor, providing an explicit prior of reflection dominance for each image and enabling image-dependent LoRA modulation. Second, we introduce a time-conditioned mechanism that fuses this reflection descriptor with timestep embeddings, enabling LoRA modulation to evolve consistently with the coarse-to-fine denoising process. By jointly modeling reflection strength and denoising dynamics, our ReLo-IRR achieves robust suppression of diverse reflection conditions. Extensive experiments on challenging benchmarks validate the effectiveness of ReLo-IRR, demonstrating superior dereflection performance and robust generalization. The code is released at \url{https://github.com/KONGBAI-8080/ReLo-IRR}. 
\end{abstract}

\begin{keywords}
Reflection removal, LoRA, Adaptive control
\end{keywords}
\section{Introduction}

Single-image reflection removal (SIRR) \cite{Li2019icassp,huang2024du,zhao2024irrfeature,zhao2025method} aims to recover the clean transmission layer from a single image degraded by undesired reflections. 
It is a fundamental yet challenging problem with broad applications in photography enhancement, autonomous driving, and augmented reality.  
The task is highly ill-posed since both transmission and reflection layers are unknown and spatially entangled, making separation ambiguous from a single observation.  

With the progress of deep learning, various methods have been developed for SIRR.  
Early CNN-based approaches \cite{hu2023single,wei2019single,li2020single} focus on network architecture and task-specific loss design.  
For instance, DSRNet \cite{hu2023single} introduces a general reflection formation model with a learnable residual term, while ERRNet \cite{wei2019single} employs an alignment-invariant loss to exploit misaligned training data.  
Although effective, these models are constrained by limited model capacity and generalize poorly to diverse real-world scenarios.
Recently, large generative models \cite{rombach2022high,esser2024scaling} have been introduced for SIRR, providing stronger performance.  
Hu \emph{et al.} \cite{hu2025dereflection} adopt a ControlNet-conditioned diffusion model, while L-DiffER \cite{hong2024differ} integrates diffusion priors with language-driven refinement.  
Despite these advances, existing diffusion-based methods typically employ image-agnostic adaptation (\emph{e.g.}, fine-tuning, ControlNet), enforcing uniform suppression across input images.
In practice, however, reflection severity varies widely: strong reflections demand aggressive suppression, while mild cases require subtle correction to avoid damaging transmission details.
Such uniform treatment often leads to residual reflections in strong cases or over-smoothing when reflections are weak. 

In this paper, we propose ReLo-IRR, a novel reflection-guided LoRA framework built upon the rectified flow (RF) model to achieve robust reflection removal.  
ReLo-IRR explicitly estimates the reflection severity of input image to adaptively regulate LoRA scaling, thereby achieving image-dependent restoration strength.  
Specifically, we design a lightweight reflection-degree estimator that predicts the descriptor of reflection dominance, which directly modulates LoRA updates.  
Furthermore, considering that RF denoising proceeds in a coarse-to-fine manner, we incorporate a timestep-conditioned mechanism that fuses reflection severity with timestep embeddings, allowing LoRA modulation to evolve consistently across the denoising process. 
We validate ReLo-IRR on challenging benchmarks, demonstrating clear improvements over state-of-the-art methods.  
Our contributions can be summarized as follows:
\begin{enumerate}
    \item We propose ReLo-IRR, a robust reflection-guided LoRA framework that explicitly incorporates reflection priors and time conditions into rectified flow inference for single-image reflection removal.  
    \item We design a lightweight reflection guidance estimator and a time-conditioned mechanism, enabling adaptive LoRA scaling across images and denoising timesteps.  
    \item Extensive experiments show that ReLo-IRR achieves state-of-the-art performance and robust generalization under varying reflection conditions.  
\end{enumerate}

\section{Related Work}

\subsection{Single-Image Reflection Removal}

Single-image reflection removal (SIRR) aims to decompose a reflection-contaminated image into transmission and reflection layers.  
Early approaches \cite{tan2005separating,yang2014efficient,shen2008chromaticity} rely on handcrafted priors and optimization techniques, such as chromatic analysis or least-squares formulations. However, these methods often fail under complex reflection conditions.  

With the advent of deep learning, CNN-based methods \cite{wei2019single,wen2019single,li2020single} achieve substantial improvements by introducing task-specific architectures and loss functions.
For example, ERRNet \cite{wei2019single} leverages alignment-invariant losses to cope with misaligned training data, and Zhu \emph{et al.} \cite{zhu2024revisiting} employ a cascade network to decouple reflection detection and removal.
Recently, diffusion-based frameworks \cite{hong2024differ,hu2025dereflection} have been explored for SIRR. 
For instance, DRR \cite{hu2025dereflection} leverages ControlNet \cite{zhang2023adding} to encode the reflection-corrupted input and employs a decoder to enhance detail preservation.  

Despite these advances, existing diffusion-based methods adopt uniform removal strategies, neglecting the varying reflection strength across scenes. 
This leads to under-removal in strong-reflection cases and over-suppression when reflections are weak.

\subsection{Adaptation and Control of Diffusion Models}

Diffusion \cite{rombach2022high} and rectified flow (RF) \cite{esser2024scaling} models have demonstrated remarkable capability in image generation tasks. 
For downstream adaptation, full finetuning of large diffusion models is straightforward but computationally expensive.  
To enhance efficiency, parameter-efficient strategies are developed. 
ControlNet \cite{zhang2023adding} augments diffusion models with task-specific conditional branches, enabling explicit guidance from external signals, such as edge and depth maps. 
Adapter-based methods \cite{mou2024t2i,ye2023ip} instead insert lightweight adapter modules into pretrained diffusion backbones, allowing flexible conditioning.
LoRA \cite{hu2022lora} injects low-rank adapters into model weights, allowing lightweight adaptation with minimal overhead. 
Subsequent extensions \cite{kopiczko2024vera,pan2024lisa,kim-etal-2024-ra} further refine LoRA adaptation. 
For example, LISA \cite{pan2024lisa} assigns different importance to layers and freezes most intermediate ones during training.
\section{Method}

\begin{figure}[t] 
	\centering
	\includegraphics[width=0.95\columnwidth]{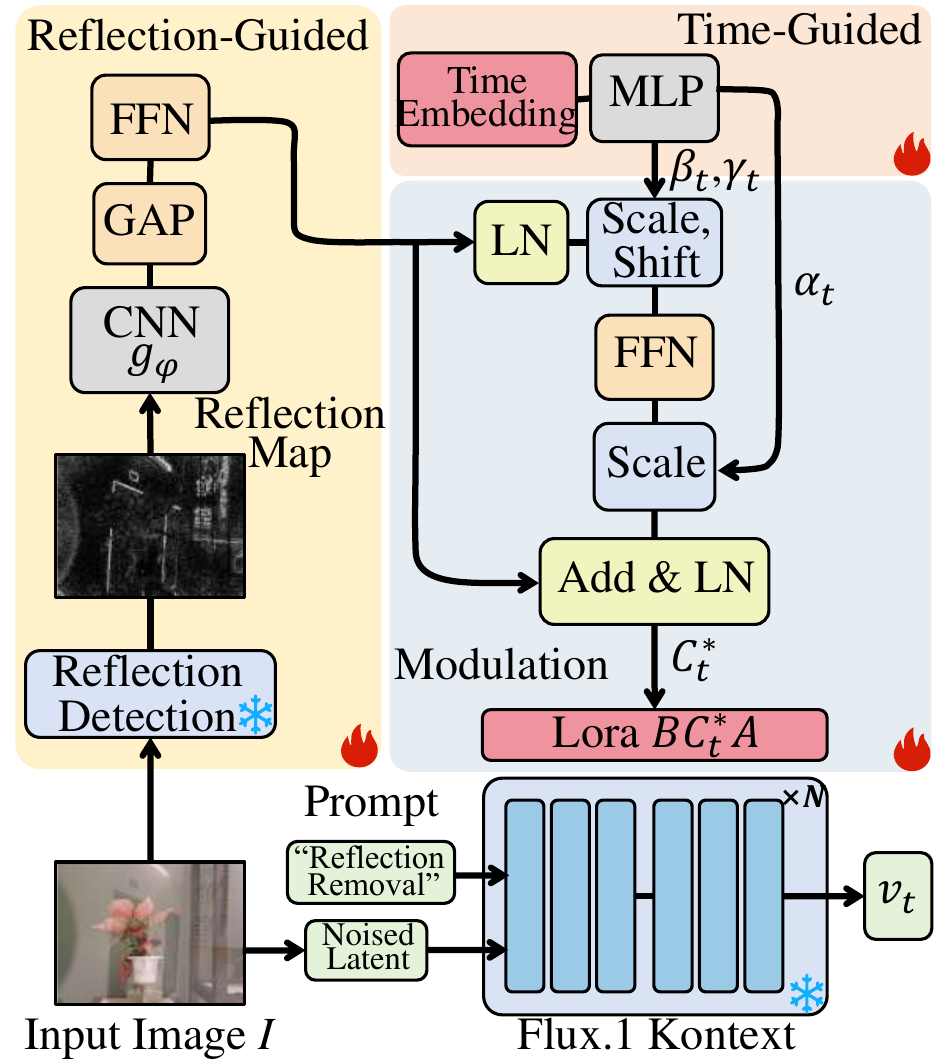}
	\caption{Overall pipeline of the proposed ReLo-IRR. Built upon Flux.1 Kontext, ReLo-IRR integrates reflection-guided and time-conditioned modulation to adapt LoRA updates.} 
	\vspace{-0.5cm} 
	\label{fig:pipeline}
\end{figure}

Given a reflection-contaminated image $I$ of resolution $H \times W$, single-image reflection removal (SIRR) aims to recover the transmission layer $\mathbf{x}$ while suppressing the reflection layer $\mathbf{r}$.  
This task is ill-posed, since both layers are unknown and spatially entangled, leading to severe ambiguities between content and artifact.  
Recent diffusion-based approaches exploit pretrained generative models, yet typically apply image-agnostic adaptation, \emph{e.g.}, a single fine-tuning or controller for all inputs, which ignores that reflection severity varies across images.  
Consequently, uniform adaptation may under-remove strong reflections or over-smooth weakly reflected images.

To address this issue, we propose ReLo-IRR, a reflection-guided LoRA framework built upon a pretrained rectified flow (RF) model $f_\theta$.  
As illustrated in Fig.~\ref{fig:pipeline}, ReLo-IRR controls LoRA updates with two signals: a reflection-guided descriptor derived from the input image, and a time-conditioned modulation that adapts the descriptor along the denoising trajectory.

\subsection{Reflection-Guided LoRA Modulation}

We first design a reflection-guided LoRA mechanism that leverages reflection priors extracted from the input image to modulate the LoRA update.  

A pretrained reflection detection network $\mathcal{R}_\psi$ \cite{zhu2024revisiting} produces a soft reflection map:
\begin{equation}
\mathbf{M} = \mathcal{R}_\psi(I), \qquad \mathbf{M}\in[0,1]^{H\times W},
\end{equation}
where larger values indicate higher probability of reflection. 
This map provides approximate spatial cues of contaminated regions and serves as a prior for estimating reflection severity.  

To transform $\mathbf{M}$ into a compact representation, a lightweight convolutional projector $g_\varphi$ is applied, followed by global average pooling and a feed-forward network:
\begin{equation}
\mathbf{h} = \mathrm{FFN}\!\left(\mathrm{GAP}(g_\varphi(\mathbf{M}))\right) \in \mathbb{R}^{1 \times d_h},
\label{eq:h}
\end{equation}
where $d_h = r^2$ denotes the dimension of the compact descriptor, with $r$ being the LoRA rank.  
We then normalize the descriptor:
\begin{equation}
\mathbf{h}_{\text{norm}} = \mathrm{LN}(\mathbf{h}) \in \mathbb{R}^{1 \times d_h}.
\end{equation}

Next, $\mathbf{h}_{\text{norm}}$ is refined through a residual feed-forward block and reshaped into a matrix form:
\begin{equation}
C = \mathrm{Reshape}\!\left(\mathrm{LN}(\mathbf{h} + \mathrm{FFN}(\mathbf{h}_{\text{norm}})), r, r\right) \in \mathbb{R}^{r \times r},
\end{equation}
where $C$ serves as the reflection-guided coefficient, summarizing global reflection characteristics.  

For a weight $W \in \mathbb{R}^{d\times n}$ in the RF model, standard LoRA~\cite{hu2022lora} introduces a low-rank update:
\begin{equation}
W^\star = W + BA, \qquad B\in\mathbb{R}^{d\times r},~ A\in\mathbb{R}^{r\times n},
\end{equation}
with $r \ll \min(d,n)$.  
Our ReLo-IRR extends this formulation by injecting the reflection-guided coefficient $C$:
\begin{equation}
W^\star = W + BC A.
\end{equation}

This design enables input-dependent LoRA scaling, \emph{i.e.,} stronger updates are applied to heavily contaminated images, while weaker updates are assigned to cleaner cases.  
As a result, ReLo-IRR effectively adapts to diverse reflection conditions and achieves robust suppression without over-smoothing.

\subsection{Time-Conditioned Reflection Guidance}

While reflection-guided LoRA modulation enables input-dependent adaptation for diverse reflection conditions, it does not account for temporal dynamics of the RF process.  
RF models denoise progressively across timesteps, \emph{i.e.,} early stages operate under heavy noise and demand stronger reflection suppression, while later stages refine structural details and benefit from more conservative modulation.  
Thus, applying a fixed suppression strength throughout all timesteps remains suboptimal, motivating a time-conditioned strategy.

\begin{table*}[t]
	\centering
	\small
    
    \caption{Quantitative results on image reflection removal. 
    The best result is bolded, and the second best is underlined.}
    \resizebox{\textwidth}{!}{  
    \begin{tabular}{c||c|c|c|c||c|c|c|c||c|c|c|c}
        \hline
        \multirow{2}{*}{Method}  & \multicolumn{4}{c||}{Nature} & \multicolumn{4}{c||}{Real}  & \multicolumn{4}{c}{RR4K}\\ \cline{2-13}
        &PSNR $\uparrow$& SSIM $\uparrow$&LPIPS $\downarrow$& DISTS $\downarrow$&PSNR $\uparrow$& SSIM $\uparrow$&LPIPS $\downarrow$& DISTS $\downarrow$&PSNR $\uparrow$& SSIM $\uparrow$&LPIPS $\downarrow$& DISTS $\downarrow$\\ \hline \hline

     ERRNet\cite{wei2019single} & 20.56 & 0.784 & 0.170 & 0.109 & 23.10 & 0.832 & 0.152 & 0.107 & 23.89 & 0.882 & 0.134 & 0.077\\ \hline
     DSRNet\cite{hu2023single} & 25.08 & 0.839 & 0.113 & 0.076  & \underline{23.85} & 0.827 & 0.146 & \underline{0.105} & 25.14 & 0.882 & 0.139 & 0.076\\ \hline
     RAGNet\cite{yu2023RAGNet} & 20.50 & 0.779 & 0.171 & 0.111  & 21.07 & 0.790 & 0.199 & 0.131 & 23.76 & 0.866 & 0.152 & 0.088\\ \hline
     RDNetRRNet\cite{zhu2024revisiting} & 26.05 & 0.846 & \textbf{0.100} & 0.075 & 21.82 & 0.802 & 0.177 & 0.120 & 25.34 & \textbf{0.891} & 0.129 & 0.078\\ \hline
     RDNet\cite{zhao2025RDNet} & \underline{26.31} & \underline{0.846} & \underline{0.103} & \underline{0.069} & \textbf{25.58} & \underline{0.846} & \textbf{0.110} & \textbf{0.083}& \underline{26.49} & \underline{0.890} & \underline{0.089} & \underline{0.069} \\ \hline \hline
     ReLo-IRR (Ours) & \textbf{27.30} & \textbf{0.860} & 0.107 & \textbf{0.064} & \textbf{25.58} & \textbf{0.855} & \underline{0.137} & \textbf{0.083} & \textbf{28.18} & 0.879 & \textbf{0.082} & \textbf{0.057}\\ \hline
     
    \end{tabular}}

	\label{tab:sota_result}
\end{table*}

\begin{figure*}[t]
	\centering
	\includegraphics[width=\linewidth]{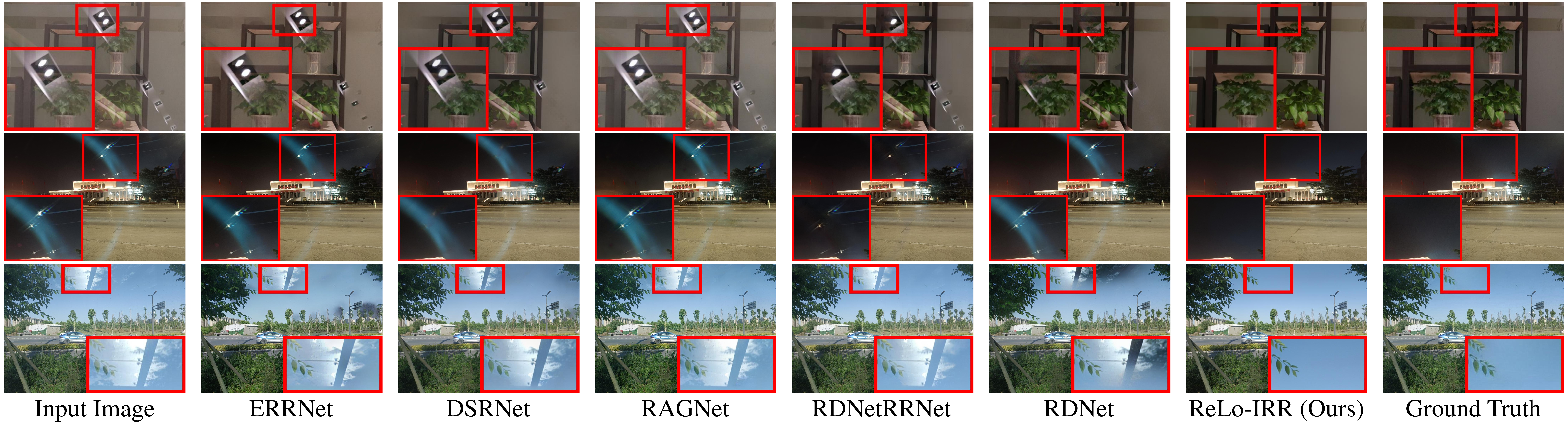}
	\caption{Qualitative comparisons on image reflection removal. Our ReLo-IRR achieves robust dereflectoion.}
		\vspace{-0.5cm} 
	\label{fig:qualitative_results}
\end{figure*}

\begin{figure}[t]
	\centering
	\includegraphics[width=1\columnwidth]{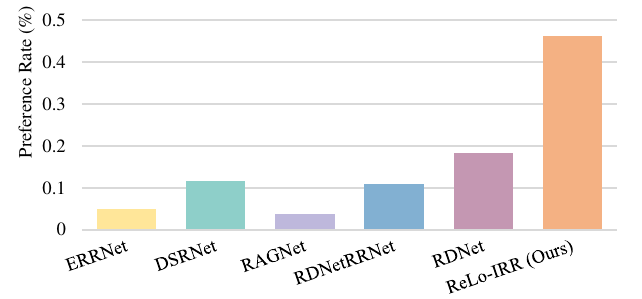}
	\caption{User study results. Our ReLo-IRR achieves the highest preference rate.}
		\vspace{-0.5cm} 
	\label{fig:userstudy}
\end{figure}

To incorporate temporal awareness, we embed the denoising timestep $t$ using sinusoidal positional encoding:
\begin{equation}
E_{(t,2i)} = \sin\!\left(\tfrac{t}{10000^{2i/D}}\right), \quad
E_{(t,2i+1)} = \cos\!\left(\tfrac{t}{10000^{2i/D}}\right),
\end{equation}
where $D$ is the embedding dimension and $i$ indexes its components.  

Following the adaptive layer normalization paradigm \cite{peebles2023dit}, the embedding $E$ is passed through an MLP $\Phi$, yielding timestep-dependent modulation parameters:
\begin{equation}
(\beta_t,\gamma_t,\alpha_t) = \Phi(E_t).
\end{equation}
These parameters scale and shift the hidden activations $\mathbf{h}$ from Eq.~\eqref{eq:h}, and adjust the reflection descriptor over timesteps to guide LoRA updates in the RF model.
First, $\mathbf{h}$ is adapted via timestep-dependent normalization:
\begin{equation}
\hat{\mathbf{h}}_{\text{norm}} = \mathrm{LN}(\mathbf{h}) \cdot (1+\gamma_t) + \beta_t.
\end{equation}
Next, the reflection descriptor is temporally modulated to align with the denoising stage:
\begin{equation}
C^\star_t = \mathrm{Reshape}\!\left(\mathrm{LN}(\mathbf{h} + \alpha_t \cdot \mathrm{FFN}(\hat{\mathbf{h}}_{\text{norm}})), r, r\right).
\end{equation}
Finally, LoRA weights are updated as:
\begin{equation}
W^\star = W + B C^\star_t A.
\end{equation}

By integrating reflection priors with timestep-aware conditioning, ReLo-IRR adaptively regulates both the magnitude and timing of reflection suppression.  
This dual guidance mechanism enhances robustness under varying contamination levels and preserves transmission details throughout the denoising trajectory, yielding a principled and interpretable framework for SIRR. 
\section{Experimental Results}

\subsection{Experimental Settings}

\noindent \textbf{Datasets and Evaluation Metrics.}
We evaluate our method on three challenging public datasets: Real~\cite{zhang2018real90}, Nature~\cite{li2020nature200} and RR4K \cite{rr4k}. 
Real consists of $110$ paired images of natural scenes, split into $90$ for training and $20$ for testing.  
Nature contains $220$ real-world image pairs captured with a Canon camera, with $200$ for training and $20$ for testing.  
RR4K offers $1326$ pairs of high-resolution $4$K images, where $1230$ are used for training and $96$ for testing.  
For evaluation, we employ four widely-used metrics: peak signal-to-noise ratio (PSNR), structural similarity index (SSIM), learned perceptual image patch similarity (LPIPS), and deep image structure and texture similarity (DISTS).



%

\noindent \textbf{Implementation Details.}
For training, data augmentation includes random rotations, horizontal flipping, and scaling.  
Flux.1 Kontext~\cite{labs2025flux} serves as the backbone. 
LoRA rank is set to $16$. LoRA matrix $A$ is initialized from a Gaussian distribution, while matrix $B$ is initialized to zeros.  
We adopt the Adam optimizer with $\beta_1=0.9$ and $\beta_2=0.999$, while the initial learning rate is set to $1 \times 10^{-4}$.    
All experiments are conducted for $500$ epochs on 8 NVIDIA A100 GPUs, with a batch size of 7 per GPU.


\subsection{Comparison with State-of-the-Art Methods}

\noindent \textbf{Single-Image Reflection Removal.}
We compare ReLo-IRR with recent state-of-the-art methods for single-image reflection removal. 
Quantitative results are reported in Table~\ref{tab:sota_result}. 
As shown, ReLo-IRR attains the best performance on most metrics and remains competitive on the others across all three datasets.
This highlights the effectiveness of integrating reflection-aware and time-conditioned adaptation.  
By dynamically modulating LoRA module according to both reflection severity and denoising stage, ReLo-IRR leverages the RF backbone more effectively and achieves superior dereflection accuracy.  
Qualitative comparisons in Fig.~\ref{fig:qualitative_results} further support these findings.  
Compared with existing methods, ReLo-IRR generates clearer transmission layers while better preserving structural details and textures with adaptive reflection suppression.  
Overall, ReLo-IRR establishes a principled and efficient framework that generalizes robustly across diverse scenes and delivers SOTA dereflection quality.

\noindent \textbf{User Study.}
While quantitative metrics provide useful indications of performance, they cannot fully capture perceptual quality.
Here, we conduct a user study comparing ReLo-IRR with state-of-the-art methods.
A total of $39$ participants were shown input images with reflections alongside dereflection results from different methods, presented in randomized order. For each set, participants were asked to select the most visually satisfactory result, with a ``Not Sure'' option provided to avoid forced choices.  
Each participant evaluated approximately $20$ image sets.
As shown in Fig.~\ref{fig:userstudy}, our ReLo-IRR received the highest preference, indicating its superior perceptual quality and robustness in diverse reflection scenarios.

\begin{table}[t]
	\centering
    
    \caption{Effect of each component of ReLo-IRR.}
    \resizebox{0.45\textwidth}{!}{  
    \begin{tabular}{c|c|c||c|c|c|c}
        \hline
       \multirow{2}{*}{LoRA} & \multicolumn{2}{c||}{Guidance}  & \multicolumn{4}{c}{Nature}   \\ \cline{2-7}
     &Reflection &Timestep   &PSNR $\uparrow$& SSIM $\uparrow$&LPIPS $\downarrow$& DISTS $\downarrow$\\ \hline \hline
        & & & 24.41 & 0.808 & 0.146 & 0.096\\ \hline
        \cmark & &   & 25.82 & 0.819 & 0.138 & 0.072  \\ \hline
     \cmark&  \cmark & & \underline{27.03} & \underline{0.836} & \underline{0.120} & \textbf{0.063}\\ \hline
     \cmark & \cmark &\cmark & \textbf{27.30} & \textbf{0.860} & \textbf{0.107} & \underline{0.064} \\ \hline
  
    \end{tabular}
    } 

	\label{tab:ablation_lora}
\end{table}

    
  


\begin{figure}[t]
	\centering
	\includegraphics[width=1\columnwidth]{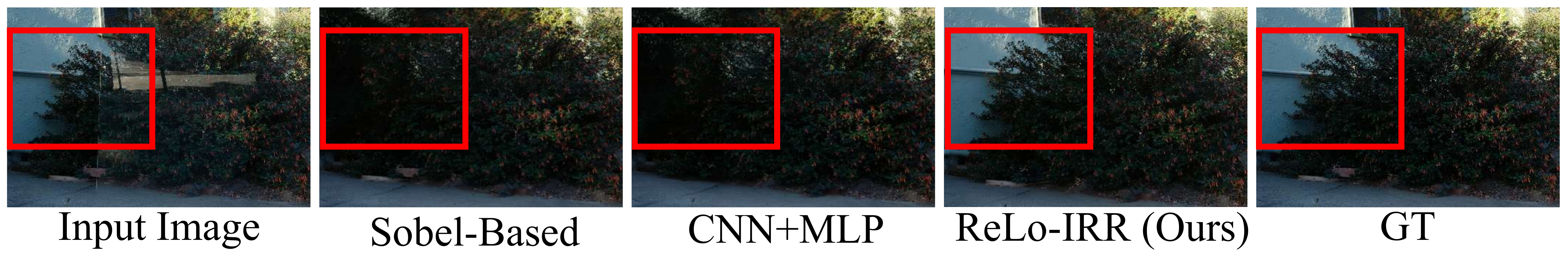}
    \caption{Effect of reflection-guidance estimator.}
		\vspace{-0.5cm} 
	\label{fig:vis_ablations2}
\end{figure}

\begin{figure}[t]
	\centering
	\includegraphics[width=1\columnwidth]{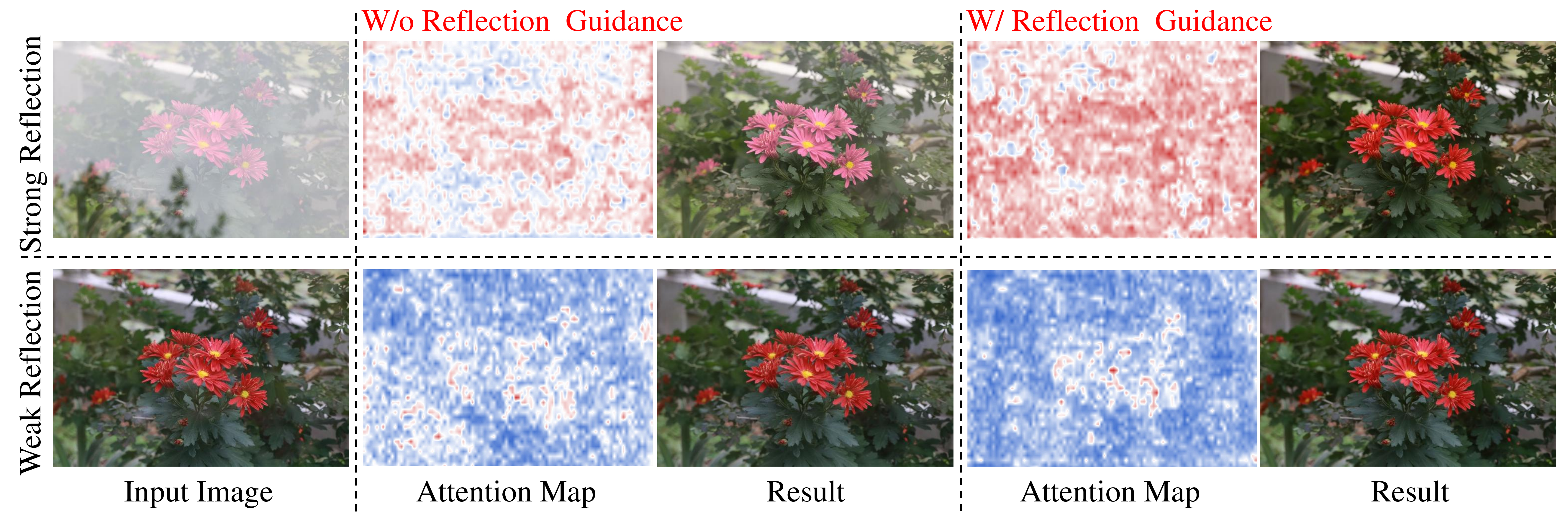}
	\caption{Visualization of attention maps of Flux and dereflection results with and without reflection-guided modulation.}
	\label{fig:vis_reflection}
\end{figure}

\begin{figure}[t]
	\centering
	\includegraphics[width=1\columnwidth]{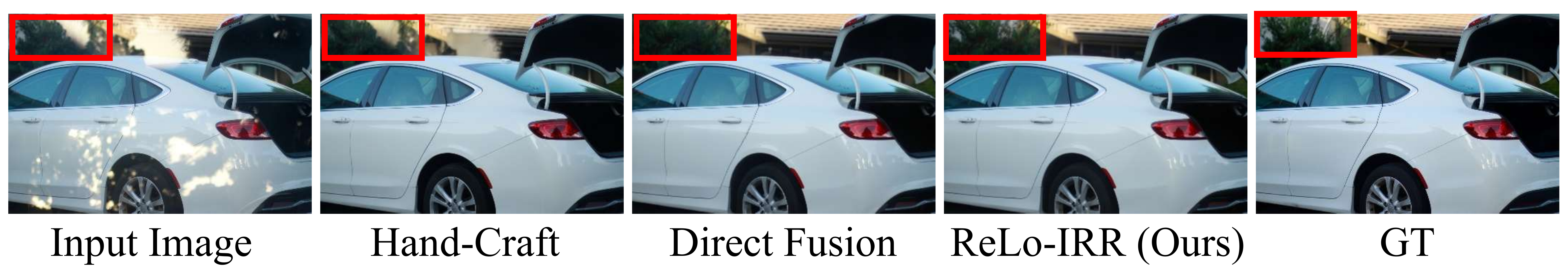}
    \caption{Effect of time-conditioned modulation.}
		\vspace{-0.3cm} 
	\label{fig:vis_ablations3}
\end{figure}

\subsection{Ablation Study}
\noindent \textbf{Effect of Reflection- and Time-Guided LoRA Control.}
The core of ReLo-IRR lies in modulating LoRA updates with both reflection- and time-guided signals. 
We assess the contribution of each component. 
As shown in Table~\ref{tab:ablation_lora}, the pretrained Flux.1 Kontext without LoRA serves as the baseline. 
Introducing LoRA fine-tuning improves dereflection quality, confirming that low-rank adaptation facilitates task specialization. 
Adding reflection guidance further boosts performance by injecting input-dependent priors that capture contamination severity. 
Further incorporating time conditioning achieves the best results, as modulation strength is adaptively aligned with the denoising trajectory. 
These results highlight the complementary roles of reflection- and time-guided strategies: the former adapts to input-specific contamination, while the latter adjusts suppression across timesteps for robust and balanced dereflection.

\noindent \textbf{Effect of Reflection-Guidance Estimator.}
ReLo-IRR adopts a reflection detection network~\cite{zhu2024revisiting} to predict a soft reflection map, which is then compressed into a global descriptor for LoRA modulation.
We compare two alternatives: 1) replacing the reflection detection network with a Sobel-based edge map followed by CNN+MLP, and 2) directly applying CNN+MLP on the input image without explicit reflection estimation.
As shown in Fig.~\ref{fig:vis_ablations2}, both designs significantly underperform.
The Sobel-based variant introduces noisy priors that confuse reflections with object boundaries, while the direct CNN variant lacks explicit reflection awareness and fails to measure contamination severity.
By contrast, our design delivers clean and reliable reflection cues, enabling the descriptor to drive adaptive and robust LoRA scaling.

\noindent \textbf{Visualization of Reflection-Guided Modulation.}
To assess the role of reflection guidance, we visualize attention maps from Flux.1 Kontext with and without reflection-guided modulation.
As shown in Fig.~\ref{fig:vis_reflection}, incorporating reflection priors makes the model more sensitive to reflection severity: it allocates stronger attention to heavily contaminated regions while avoiding redundant corrections for the cleaner image.
This validates that reflection-guided LoRA enables selective adaptation, \emph{i.e.,} intensifying updates only when necessary and preserving structural details otherwise.

\noindent \textbf{Effect of Time-Conditioned Modulation.}
We inject timestep information through AdaNorm, where temporal embeddings generate scale-shift parameters to modulate normalized features.  
We compare this design with two alternatives: 1) a hand-crafted schedule, where modulation strength is initialized set to $1.2$ and decays to $1$ after $10$ steps, and 2) direct fusion of timestep embeddings with reflection priors via an MLP.  
As shown in Fig.~\ref{fig:vis_ablations3}, our AdaNorm delivers the best performance.  
While the fixed schedule lacks adaptivity and the MLP fusion introduces unstable modulation, AdaNorm achieves smooth and stage-aware control, enabling adaptive suppression along the denoising trajectory.

\section{Conclusion}
In this paper, we propose ReLo-IRR, a novel reflection-guided LoRA framework for single-image reflection removal.  
ReLo-IRR explicitly estimates the reflection severity and incorporates it into lightweight LoRA adapters, enabling image-dependent dereflection that adapts to varying reflection conditions.  
By further integrating a timestep-conditioned mechanism aligned with the coarse-to-fine dynamics of rectified flow models, ReLo-IRR modulates restoration consistently across the denoising.  
Extensive experiments demonstrate that ReLo-IRR achieves state-of-the-art performance.

\clearpage

\section{Acknowledgments}
This work was supported by Guangdong Basic and Applied Basic Research Foundation under grants 2024A1515140109 and 2023A1515110695.

\bibliographystyle{IEEEbib}
\bibliography{strings,refs}

\end{document}